\documentclass[conference]{IEEEtran}
\IEEEoverridecommandlockouts
\usepackage[T1]{fontenc}
\usepackage[utf8]{inputenc}
\usepackage{amsmath,amssymb,amsfonts}
\usepackage{algorithm}
\usepackage{algpseudocode}
\usepackage{graphicx}
\usepackage{textcomp}
\usepackage{xcolor}
    
\begin{document}

\title{Synthetic Dataset Creation and Fine-Tuning of Transformer Models for Question Answering in Serbian}

\author{
\IEEEauthorblockN{Aleksa Cvetanović}
\IEEEauthorblockA{\textit{School of Electrical Engineering} \\
\textit{University in Belgrade}\\
Belgrade, Serbia \\
aleksa.cvetanovic99@gmail.com}
\and
\IEEEauthorblockN{Predrag Tadić}
\IEEEauthorblockA{\textit{School of Electrical Engineering} \\
\textit{University in Belgrade}\\
Belgrade, Serbia \\
ptadic@etf.bg.ac.rs}
}

\maketitle

\begin{abstract}
In this paper, we focus on generating a synthetic question answering (QA) dataset using an adapted Translate-Align-Retrieve method. Using this method, we created the largest Serbian QA dataset of more than 87K samples, which we name SQuAD-sr. To acknowledge the script duality in Serbian, we generated both Cyrillic and Latin versions of the dataset. We investigate the dataset quality and use it to fine-tune several pre-trained QA models.
Best results were obtained by fine-tuning the BERTić model on our Latin SQuAD-sr dataset, achieving 73.91\% Exact Match and 82.97\% F1 score on the benchmark XQuAD dataset, which we translated into Serbian for the purpose of evaluation.
The results show that our model exceeds zero-shot baselines, but fails to go beyond human performance. We note the advantage of using a monolingual pre-trained model over multilingual, as well as the performance increase gained by using Latin over Cyrillic. By performing additional analysis, we show that questions about numeric values or dates are more likely to be answered correctly than other types of questions. Finally, we conclude that SQuAD-sr is of sufficient quality for fine-tuning a
Serbian QA model, in the absence of a manually crafted and annotated dataset.
\end{abstract}


\section{Introduction}

Extractive Question Answering (usually referred to as Question Answering or QA) is one of the most common tasks in Natural Language Processing (NLP). It assesses the Reading Comprehension (RC) ability of a machine by setting an objective to extract the required information from a text provided in natural language. Given the context and question, the goal is to extract a span from the context which represents the best possible answer to the posed question. Contrary to Generative Question Answering, Extractive Question Answering performs the extraction of the answer without changing word order and form in the extracted span.

In recent years there has been enormous progress in every aspect of NLP. With the introduction of Transformers \cite{DBLP:journals/corr/VaswaniSPUJGKP17},
many 
models which compare to or even exceed human performance have been released. By introducing new fine-tuning techniques and scaling model size up to more than a trillion parameters \cite{openai2023gpt4}, current models act as an all-in-one solution to all NLP tasks. However, due to the computational requirements for 
these Large Language Models (LLM), smaller models that focus on a single task remain relevant to this day. In this paper we 
focus on Transformer-based models with 
parameter count ranging from 110M \cite{DBLP:journals/corr/abs-2104-09243} to 270M \cite{DBLP:journals/corr/abs-1911-02116}, enabling the possibility to train and run these models without supercomputers.

Due to its omnipresence in both the real and digital world, English is used as the default language in NLP research. Except for other languages with large speaker populations such as Chinese or French, there is a substantial gap in model performance on low-resource languages, regardless of the specific NLP task. Additionally, aside from unlabeled natural text, there are little to no datasets suitable for fine-tuning a model for a specific task. The absence of labeled high-quality data aggravates the process of releasing an NLP model for low-resource languages.

To acquire a dataset comparable to existing English ones in terms of both size and quality, a lot of investment is required. Hiring crowd workers to manually label the data takes time and results in significant costs. For those reasons, the idea is posed to craft a synthetic dataset based on already existing ones.

In this paper, we fine-tune a QA model in Serbian. To overcome the issue of data shortage, we turn to synthetic dataset generation methods. We follow the idea of the Translate-Align-Retrieve method from \cite{DBLP:journals/corr/abs-1912-05200} to which we propose changes to make it more suitable for Serbian. By applying the method to SQuAD v1.1 \cite{rajpurkar2016squad}, we obtain a dataset which we utilize to fine-tune several models. We 
analyse the dataset and compare the performance of acquired models on the XQuAD \cite{DBLP:journals/corr/abs-1910-11856} dataset 
which we translated to Serbian.

The contributions of this work are as follows:
\begin{itemize}
    \item We publish the largest question answering dataset in Serbian\footnote{https://www.kaggle.com/datasets/aleksacvetanovic/squad-sr}
    \item We release the best-performing model acquired from our research\footnote{https://huggingface.co/aleksahet/BERTic-squad-sr-lat}
    \item We perform extensive analysis on both the dataset and the fine-tuning design choices and report the conclusions
\end{itemize}

\section{Related work}

Undoubtedly, English is the 
language with the most available resources for extractive question answering. The most prominent question answering dataset is the Stanford Question Answering Dataset (SQuAD) v1.1 \cite{rajpurkar2016squad}, often used as a benchmark for pre-trained models. Datasets such as NewsQA \cite{trischler2017newsqa} and Natural Questions \cite{kwiatkowski-etal-2019-natural} follow a similar structure and open domain property, as opposed to domain-specific datasets such as SubjQA \cite{DBLP:journals/corr/abs-2004-14283}. There are also monolingual datasets for some of the other languages with large speaker communities---SberBERT \cite{DBLP:journals/corr/abs-1912-09723} in Russian or FQuAD \cite{DBLP:journals/corr/abs-2002-06071} in French---which have all been collected by employing crowd workers.

The advancement of available solutions has prompted researchers to make the QA task more difficult, by adding samples with unanswerable questions to the datasets. An example is SQuAD v2.0 \cite{rajpurkar2018know}, which contains over 50K unanswerable questions. However, in this research we focus on the simpler version of the problem, leaving unanswerable questions handling as future work.

Some attempts to address the issue of data scarcity in languages other than English have been made by crafting multilingual question answering datasets, such as MLQA \cite{DBLP:journals/corr/abs-1910-07475} or XQuAD \cite{DBLP:journals/corr/abs-1910-11856}. These datasets are intended to be used primarily for assessing zero-shot cross-lingual question answering, but due to the lack of other datasets, some researchers evaluate their solutions on portions of these datasets that contain samples in the required language. Since these datasets cover only a handful of languages---neither of which is Serbian---we manually translate XQuAD to Serbian and use it to evaluate our models.

Other researchers have also explored the idea of crafting a synthetic dataset to tackle the data scarcity issue. In \cite{DBLP:journals/corr/abs-1906-05394}, the authors use only samples in which answer translation can be matched to a span of a context translation. Mozannar et al.\ \cite{lee-etal-2018-semi} utilize synthetic dataset to train a weak QA system, and manually crafted dataset for further fine-tuning. In \cite{DBLP:journals/corr/abs-1906-05394}, edit distance is utilized to find context spans that correspond to translated answers. Finally, \cite{DBLP:journals/corr/abs-1912-05200} proposes Translate-Align-Retrieve (TAR) method which utilizes word alignments to extract answer span from the context. We recognize the TAR method as the most suitable for Serbian, due to frequent form changes in the answer translation, and further adapt it to increase the synthetic samples quality.

To the best of our knowledge, the QA task in Serbian hasn't been tackled directly. However, besides the zero-shot approach, we can use multi-purpose Large Language Models (LLMs) such as Llama-2 \cite{touvron2023llama}. Although these models prove to be somewhat effective, their size makes them unusable on the average computer. In this paper, we turn to the approach of fine-tuning a BERT-like pre-trained model to acquire an effective Serbian QA system.

Regarding pre-trained models, several multilingual variants have been released in previous research \cite{DBLP:journals/corr/abs-1810-04805, DBLP:journals/corr/abs-1911-02116}. Also, a BERT variant trained on Serbo-Croatian macro-language \cite{DBLP:journals/corr/abs-2104-09243} named BERTić has been released publicly. We test the impact of the pre-trained model by fine-tuning both monolingual and multilingual variants and report the results.

\section{Methodology}

Our work is split into two main parts: dataset synthesis and model fine-tuning. By performing a method described in Chapter \ref{synthesis} we acquire the largest Serbian QA dataset, which we name SQuAD-sr. Addressing the script duality property of Serbian, we acquire the dataset in both Cyrillic and Latin. Fine-tuning provides us with several models which we evaluate and report the results in Chapter \ref{results}.

\subsection{Dataset Synthesis} \label{synthesis}

To obtain the synthetic dataset, we start with the data provided in SQuAD v1.1. This is a publicly available QA dataset with more than 87K high-quality samples in its train split. The data is crowdsourced by having manual workers provide questions and answer spans from contexts scraped from Wikipedia. The dataset is often used \cite{ DBLP:journals/corr/abs-1911-02116, DBLP:journals/corr/abs-1810-04805, liu2019roberta} for fine-tuning and evaluation of pre-trained models, due to its quality, size, and variety of topics included. 

Using the Translate-Transliterate-Align-Retrieve method inspired by \cite{DBLP:journals/corr/abs-1912-05200}, we convert each triple \((c_{en}, q_{en}, a_{en})\) from English to Serbian \((c_{sr}, q_{sr}, a_{sr})\), where \(c, q\) and \(a\) represent context, question and answer, respectively.

\subsubsection{Translate}

Before the translation, we split each context into single sentences to improve the translation quality. We use Punkt sentence tokenizer implementation in NLTK toolkit \cite{bird2009natural}. Sentences are then processed by trimming whitespaces. We translate all article titles, contexts, and questions from the dataset. It's important to note that answers acquired by translating \(a_{en}\) are unsuitable because answers are required to keep the same form as in the given context. In later steps, we extract answers from \(c_{sr}\).

To perform the translation, we utilize the publicly available Neural Machine Translation (NMT) model NLLB-200-1.3B from Meta AI \cite{nllbteam2022language}. When translating to Serbian, the model outputs sentences in Cyrillic, which is suspected to affect the model quality. To address this issue, we introduce a Transliterate step, which we mark as optional.

\subsubsection{Transliterate}

Fine-tuning a model on Cyrillic text might negatively impact the results because pre-trained models are usually trained on a much larger portion of text in Latin. Since in Serbian Cyrillic and Latin bear equal status, we introduce Transliterate step to obtain the sentences in Latin and mark 
it as optional. In Chapter \ref{results} we investigate and discuss the impact of the script on the evaluation results. 
To perform the transliteration, we utilize the publicly available Cyrillic-transliteration module \cite{georges_labreche_2023_7734906}.

\subsubsection{Align}

To extract \(a_{sr}\) from \(c_{sr}\) we need information about word-to-word correspondence for each sentence in \(c_{en}\) and \(c_{sr}\). The task of finding such correspondence is named Word Alignment and is considered a legacy subtask of word-based Statistical Machine Translation (SMT). It consists of finding which word from a sentence in \(c_{sr}\) acts as a translation of each word from a sentence in \(c_{en}\), and vice versa.

To obtain word alignment results we utilize Efficient Low-Memory Aligner (eflomal) \cite{eflomal}, a publicly available unsupervised word alignment tool based on Gibbs sampling with a Bayesian extension of the IBM alignment models. Before applying eflomal, we create word tokens for each sentence using TreeBank tokenizer from the NLTK toolkit \cite{bird2009natural}, and therefore structure the data according to the required format.

To further improve the alignments we apply grow-diag-final-and heuristics to the eflomal results. We use the implementation from the fast-align repository \cite{DBLP:conf/naacl/DyerCS13}.

\subsubsection{Retrieve}

The final step utilizes previously acquired translations and alignments to obtain the synthetic dataset. We employ a specific strategy depending on the text sequence - context, question, or answer.

\textbf{Contexts.}
Translated contexts are obtained by concatenating corresponding individually translated sentences. Additionally, we compute mappings \(char2word_{en}\) and \(word2char_{sr}\) which map the first character index of each word to the word index, and the word index of each word to the first character index, respectively.

Finally, to extract the answer span from \(c_{sr}\) context alignments are required. These are computed from previously acquired sentence alignments of individual sentence pairs contained in \(c_{en}\) and \(c_{sr}\). For each sentence pair, we increment its sentence alignments by the total number of word tokens contained in previous sentence pairs. Naturally, we count English word tokens to increment English alignments, and Serbian word tokens to increment Serbian alignments.

\textbf{Questions.} We use question translations from the translation step without any alterations.

\textbf{Answers.} We leverage previously computed context alignments, as well as \(char2word_{en}\) and \(word2char_{sr}\) to extract the answers from the contexts. First, we use \(char2word_{en}\) to get word indices of each word contained in \(a_{en}\). Then, for each word index, we employ context alignment to determine the corresponding Serbian word index to which each word from \(a_{en}\) is mapped. If no words from \(a_{en}\) are mapped to Serbian we drop the sample. By calculating the smallest and the largest indices of Serbian words, we determine the words which are contained in \(a_{sr}\). Finally, we employ \(word2char_{sr}\) to extract the first character index of the first word.

\subsubsection{SQuAD-sr}

The method above provides us with a synthetic dataset containing 87175 examples---only 424 examples less than SQuAD v1.1. We compare it to SQuAD v1.1 and Serbian XQuAD split in terms of size and average sequence length and report the results in Table \ref{datastats}. We note the differences between average context lengths. The context length difference between SQuAD v1.1 and the Latin version of SQuAD-sr arguably stems from the differences between English and Serbian. The difference between average context lengths of Latin SQuAD-sr and Serbian XQuAD indicates longer contexts in XQuAD, and consequently harder evaluation examples.

\begin{table}[ht]
    \begin{center}
    \caption{
    Dataset metrics comparison. {\it SQuAD-sr}: Latin version of the synthetically generated dataset. {\it XQuAD}: XQuAD samples manually translated to Serbian}
    \label{datastats}
        \begin{tabular}{c c c c}
              & SQuAD v1.1 & SQuAD-sr & XQuAD \\
             \hline
             \# samples & 87599 & 87175 & 1190 \\
             context length & 736 & 715 & 758 \\
             question length & 60 & 57 & 58 \\
             answer length & 20 & 20 & 19 \\
        \end{tabular}
    \end{center}
\end{table}

To assess the dataset quality, we manually inspect a portion of SQuAD-sr. We focus on examining samples that significantly deviate from the mean in terms of context, question or answer length, as well as additional 500 random samples.

The usage of the NMT system is noticeable in both contexts and questions. In some samples translation error also propagates to the answer, making it poorly translated. By inspecting long contexts, questions, and answers, we notice that the NMT model sometimes tends to output the same token or short sequence of tokens multiple times.

The answer extraction method yields decent results. However, by inspecting samples with short answers, we notice that the method sometimes extracts punctuation marks instead of the desired word.

Despite the occasional errors, we are satisfied with the final state of SQuAD-sr and use it as is to fine-tune our models.

\subsection{Fine-tuning}

When developing NLP systems, fine-tuning a pre-trained model for downstream tasks is considered standard practice. We apply this approach by utilizing SQuAD-sr to fine-tune neural networks for question answering. We consider several Transformer-based models \cite{DBLP:journals/corr/VaswaniSPUJGKP17}, which differ in the approach and data used in pre-training. We select two multilingual models - mBERT \cite{DBLP:journals/corr/abs-1810-04805} and XLM-R \cite{DBLP:journals/corr/abs-1911-02116}, as well as BERTić \cite{DBLP:journals/corr/abs-2104-09243} which can be considered monolingual since it has been trained on the Serbo-Croatian macro-language. Fine-tuning is performed using model and training implementations in the HuggingFace ecosystem.

\section{Experimental setup}

To properly investigate the influence of design choices, we standardize the fine-tuning process. The same hyperparameters presented in Table \ref{finetuningparams} are used for each experiment. We use Nvidia Tesla P100 GPU and train each model for approximately 4 hours.

First, we compare our models to the ones we fine-tuned using the zero-shot learning approach. Then, the influence of selecting a monolingual vs.\ multilingual model is examined. Finally, we investigate how the choice of script (Cyrillic vs. Latin) affects the performance of our model.

For our metrics, we use Exact Match (EM) and F1 Score, which are the most common metrics in extractive question answering. EM is more rigorous and awards only a full match of the prediction and the ground truth, whereas F1 Score is more forgivable and allows a partial overlap. We use the implementation of the available metric from HuggingFace.

\begin{table}[ht]
    \begin{center}
    \caption{
    Model fine-tuning hyperparameters.
    }
    \label{finetuningparams}
        \begin{tabular}{c c c}
              Batch size & Learning rate & \# epochs \\
             \hline
             16 & 3e-5 & 3 \\
        \end{tabular}
    \end{center}
\end{table}

\subsection{Evaluation Dataset}

Unfortunately, there are no datasets suitable for the evaluation of Serbian QA systems. To overcome this issue, we select a multilingual dataset named XQuAD \cite{DBLP:journals/corr/abs-1910-11856}, which consists of 1190 samples extracted from SQuAD v1.1 dev split professionally translated to 12 different languages, and translate it to Serbian. We translate contexts using Google Translate and manually evaluate and correct them. Questions are manually translated without the help of any machine translation software. In our opinion, evaluating the models on this dataset provides us with reliable scores which can be used to compare the models.

\section{Results} \label{results}

The results are presented in Table \ref{tresults}. We achieve the best performance with BERTić-SQuAD-sr with 73.91 / 82.97 \% EM / F1 on the Latin version of Serbian XQuAD. On Cyrillic, the best model is XLM-R-SQuAD-sr with 67.58 / 78.08 \% EM / F1 Score. The results provide insight into the dataset quality, as well as the influence of script and pre-trained model selection.

\subsection{Comparison against zero-shot models and human baseline}

To acquire the baseline models, we fine-tune mBERT and XLM-R on SQuAD v1.1 using the same procedure as our other models. The models are evaluated on Serbian XQuAD split in Latin. mBERT and XLM-R achieve 58.60 / 61.08 \% Exact Match and 71.71 / 73.94 \% F1 Score, respectively. We note the slight improvement of XLM-R over mBERT, which is in accordance with \cite{DBLP:journals/corr/abs-1911-02116}.

All our models achieve better results than the model baselines. This confirms the quality of SQuAD-sr, in both its Cyrillic and Latin forms, and justifies the idea of using a synthetic dataset to fine-tune a QA model. The biggest improvement over mBERT and XLM-R is achieved by Latin BERTić-SQuAD-sr - 15.31 / 12.83 \% EM and 11.26 / 9.03 \% F1, respectively.

Due to the lack of human performance results in Serbian, we compare our models to English results on SQuAD v1.1 reported in \cite{rajpurkar2016squad}. Humans outperform our model by 8.39 / 8.25 \% EM / F1, leaving room for further development of Serbian QA models.

\subsection{Monolingual vs. multilingual pre-trained models}

To investigate the impact of the choice between the monolingual and multilingual pre-trained model, we fine-tune multiple models on our SQuAD-sr. For the monolingual model we select BERTić \cite{DBLP:journals/corr/abs-2104-09243}, the best-performing monolingual model in Serbian. We test it against mBERT and XLM-R.

The results show a significant performance increase when using the monolingual model fine-tuned on the Latin dataset. We achieve a 6.33 / 4.89 \% EM / F1 increase compared to the best performing multilingual model, i.e.\ XLM-R-SQuAD-sr. Similar results are shown in \cite{DBLP:journals/corr/abs-2002-06071}, confirming the importance of having a monolingual model for fine-tuning.

\subsection{Cyrillic vs. Latin script}

Finally, we show the difference in results acquired by fine-tuning models on Cyrillic and Latin datasets. We achieve performance increases in all models when fine-tuning on the Latin dataset. The main reason is arguably the fact that all considered models use a much smaller portion of Cyrillic data for pre-training, which affects both vocabulary building and representation learning stages of pre-training. The largest difference is reported on BERTić-SQuAD-sr - 18.49/17.3 \% EM/F1 increase with the Latin dataset. The reason this model performs significantly worse on Cyrillic is presumably the vocabulary size of only 32K tokens, which is in our opinion not enough to address the Cyrillic data. Other models have larger vocabulary sizes helping them to address the Cyrillic data. Additionally, they use Russian text among others for pre-training, which makes the Cyrillic portion of the data much larger and results in more learned Cyrillic tokens.

\begin{table}[ht]
    \begin{center}
    \caption{
    Evaluation results. {\it Dataset}: The dataset we use to fine-tune the model. {\it Script}: Denotes the script of both training and test datasets (C - Cyrillic, L - Latin).
    }
    \label{tresults}
    \begin{tabular}{c  c  c  c  c}
            System & Dataset & Script & EM[\%] & F1[\%] \\
            \hline
            \multicolumn{5}{c}{Baselines} \\
            \hline
            \hline
            Human \cite{rajpurkar2016squad} &  -  & L & 82.30 & 91.22 \\
            mBERT & SQuAD v1.1 & L & 58.60 & 71.71 \\
            XLM-R & SQuAD v1.1 & L & 61.08 & 73.94 \\
            \hline
            \multicolumn{5}{c}{Our models} \\
            \hline
            \hline
            mBERT-SQuAD-sr & SQuAD-sr & \begin{tabular}{@{}l@{}}
                       C \\
                       L \\
                     \end{tabular}
                &\begin{tabular}{@{}l@{}}
                       63.68 \\
                       65.78 \\
                     \end{tabular} & \begin{tabular}{@{}l@{}}
                       75.16 \\
                       77.45 \\
                     \end{tabular} \\
            XLM-R-SQuAD-sr & SQuAD-sr & \begin{tabular}{@{}l@{}}
                       C \\
                       L \\
                     \end{tabular}
            &\begin{tabular}{@{}l@{}}
                       67.15 \\
                       67.58 \\
                     \end{tabular} & \begin{tabular}{@{}l@{}}
                       77.86 \\
                       78.08 \\
                     \end{tabular} \\
            BERTic-SQuAD-sr & SQuAD-sr & \begin{tabular}{@{}l@{}}
                       C \\
                       L \\
                     \end{tabular}
            &\begin{tabular}{@{}l@{}}
                       55.42 \\
                       73.91 \\
                     \end{tabular} & \begin{tabular}{@{}l@{}}
                       65.67 \\
                       82.97 \\
                     \end{tabular} \\
        \end{tabular}
    \end{center}
\end{table}

\subsection{Analysis}

To gain better insight into the model's 
errors, we follow a similar performance analysis approach as in \cite{DBLP:journals/corr/abs-2002-06071}. We divide samples contained in Serbian XQuAD into categories based on the question type. We use seven categories: Who (\textit{Ko, Koji, Koje, Koja}), What (\textit{Šta}), How (\textit{Kako}), When (\textit{Kad, Kada}), Where (\textit{Gde}), How many (\textit{Koliko, Koliki, Kolika}) and Other. We run the experiment using BERTić-SQuAD-sr on the Latin version of the dataset and report the results in Table \ref{analysis}.

\begin{table}[ht]
    \begin{center}
    \caption{
    Evaluation results by question type. \(CL\): average context length, \(QL\): Average question length, \(PAL\): Predicted answer length, \(RAL\): Target answer length.
    }
    \label{analysis}
        \begin{tabular}{c c c c c c c}
              Question Type & EM[\%] & F1[\%] & CL & QL & PAL & RAL  \\
             \hline
             Who & 76.56 & 84.38 & 738 & 61 & 18 & 17 \\
             What & 60.56 & 73.50 & 740 & 52 & 28 & 26 \\
             How & 80.20 & 87.26 & 751 & 58 & 19 & 19 \\
             When & 83.02 & 87.45 & 744 & 59 & 11 & 11 \\
             Where & 60.53 & 76.43 & 740 & 50 & 18 & 22 \\
             How many & 86.36 & 91.24 & 790 & 58 & 9 & 9 \\
             Other & 56.82 & 75.73 & 711 & 51 & 28 & 25 \\
        \end{tabular}
    \end{center}
\end{table}

Our model achieves the best performance on questions containing words \textit{When} and \textit{How many}. This is expected, due to the question's unambiguity and well-established answer forms. \textit{When}-questions clearly refer to dates and \textit{How many}-questions require numeric value as an answer---these relations are arguably easy to learn which results in much larger scores than the average. We also note the same average answer length of predictions and targets.

On the other hand, the model performs significantly worse on \textit{What}- and \textit{Where}-questions. \textit{What}-questions are arguably more difficult than others because they require much deeper reading comprehension, resulting in lower scores. Poor performance on \textit{Where}-questions comes as a surprise at first, due to question unambiguity---they usually refer to a location. Although the form isn't as strict as in \textit{When}-questions, the difference between the scores is much larger than expected. We believe that poor performance on these questions stems from two factors: answers to these questions are usually in genitive which makes them harder to be recognized by the model, and they are usually preceded by a preposition which is not always extracted. In our opinion, these obstacles cannot be overcome with a synthetic training dataset.

We note interesting results on \textit{How}-questions, whose scores achieved on these questions are much larger than the ones reported by \cite{DBLP:journals/corr/abs-2002-06071}. This is due to the property of Serbian to pose questions using How (\textit{Kako}) to ask for a name or a location (\textit{Kako se zove...?}, \textit{Kako glasi...?}, etc.). Contrary to \textit{Where}-questions, \textit{How}-questions referring to a location don't require answers in the genitive, which results in much higher scores.

Finally, we note the positive correlation between answer length and performance---our model achieves better performance on questions that require shorter answers. This behavior is expected and justifies the usage of answer length as a question difficulty indicator.

\section{Conclusion}

In this work, we presented SQuAD-sr, the largest QA dataset in Serbian with more than 87K samples. By focusing on synthetic dataset crafting methods, we managed to acquire SQuAD-sr much faster and with almost no investments, while still maintaining a reasonable quality. We utilized it to fine-tune several QA models, acquiring the best results with BERTić-SQuAD-sr - 73.91 / 82.97 \% EM / F1 on the Latin version of Serbian XQuAD. By achieving these scores we confirmed that a synthetic dataset can be used to successfully fine-tune a QA model, due to the lack of manually crafted Serbian QA datasets. The results exceed
zero-shot baselines but fail to go beyond human performance. Our experiments showed that monolingual pre-trained models are more suitable than multilingual ones, 
and that the Latin dataset is better for fine-tuning those models. Finally, by performing a question-type classification, we gained an insight into what questions can be asked with more confidence about the answer. Finally, we make both Latin and Cyrillic versions of the dataset publicly available, as well as BERTić-SQuAD-sr, our best-performing model.

\bibliographystyle{unsrt}
\bibliography{references}

\end{document}